\documentclass[UTF8,lettersize,journal]{IEEEtran}
\usepackage{amsmath,amsfonts}
\usepackage{algorithmic}
\usepackage{algorithm}
\usepackage{array}
\usepackage[caption=false,font=normalsize,labelfont=sf,textfont=sf]{subfig}
\usepackage{textcomp}
\usepackage{stfloats}
\usepackage{url}
\usepackage{verbatim}
\usepackage{graphicx}
\usepackage{cite}
\usepackage{xcolor}
\usepackage{multirow}
\usepackage{CJKutf8}
\usepackage{makecell}
\usepackage{booktabs}
\usepackage{adjustbox}

\usepackage{stfloats}
\usepackage{xurl}
\usepackage{xspace,mfirstuc,tabulary}
\newcommand\ZhEn{Zh$\Rightarrow$En }
\newcommand\EnZh{En$\Rightarrow$Zh }

\begin{document}

\title{Towards Privacy-Preserving Machine 
Translation at the Inference Stage: A New Task and Benchmark}

\author{Wei Shao, Lemao Liu, Yinqiao Li, Guoping Huang, Shuming Shi, Linqi Song\textsuperscript{${\dagger}$}

\thanks{\textit{\textsuperscript{${\dagger}$}Corresponding author}
	}
	\IEEEcompsocitemizethanks{
		\IEEEcompsocthanksitem 

		 Wei Shao, Yinqiao Li and Linqi Song are with the City University of Hong Kong and City University of Hong Kong Shenzhen Research Institute. 
  Lemao Liu, Guoping Huang and Shuming Shi are with the Tencent AI Lab, Shenzhen. 
	}
}



\maketitle

\begin{abstract}

Current online translation services require sending user text to cloud servers, posing a risk of privacy leakage when the text contains sensitive information. This risk hinders the application of online translation services in privacy-sensitive scenarios. One way to mitigate this risk for online translation services is introducing privacy protection mechanisms targeting the inference stage of translation models. However, compared to subfields of NLP like text classification and summarization, the machine translation research community has limited exploration of privacy protection during the inference stage. There is no clearly defined privacy protection task for the inference stage, dedicated evaluation datasets and metrics, and reference benchmark methods. The absence of these elements has seriously constrained researchers' in-depth exploration of this direction. To bridge this gap, this paper proposes a novel "Privacy-Preserving Machine Translation" (PPMT) task, aiming to protect the private information in text during the model inference stage. For this task, we constructed three benchmark test datasets, designed corresponding evaluation metrics, and proposed a series of benchmark methods as a starting point for this task. The definition of privacy is complex and diverse. Considering that named entities often contain a large amount of personal privacy and commercial secrets, we have focused our research on protecting only the named entity's privacy in the text. We expect this research work will provide a new perspective and a solid foundation for the privacy protection problem in machine translation.
\end{abstract}


\section{Introduction}

\begin{figure*}
    \centering
    \includegraphics[width=0.8\linewidth]{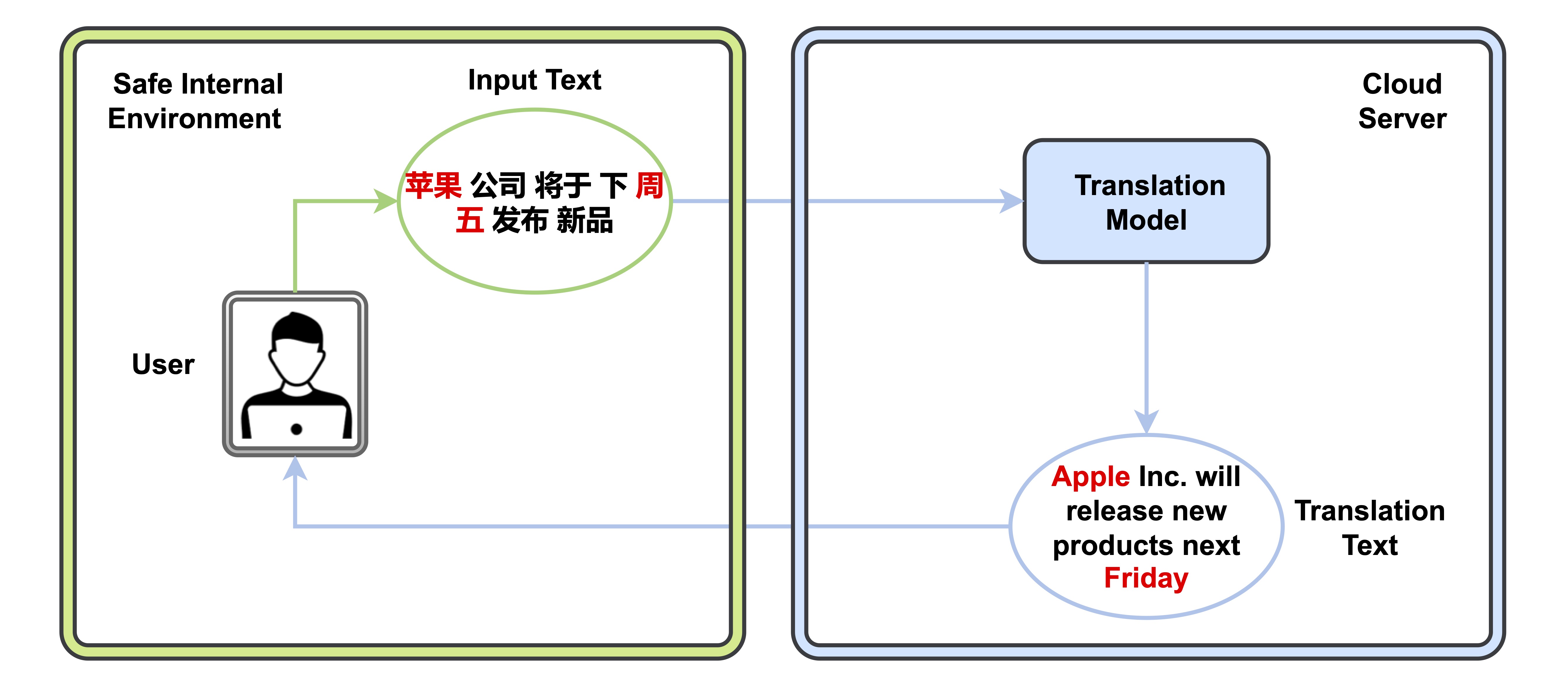}
    \caption{This figure illustrates the workflow of an online translation service. Users are required to transfer the text to be translated from a safe internal environment to the online translation system located in cloud server. The red-colored content ("苹果-Apple", "周五-Friday") in the text represents the privacy information, which constitutes sensitive business information. In sensitive scenarios, users do not want this text, which contains business information, to leave the secure internal environment (typically the company's internal network). However, the current service mode of online translation systems makes it challenging to avoid this issue.}
    \label{translation_paradigm}
\end{figure*}

With the outstanding performance of neural translation models~\cite{wu2016google, gehring2017convolutional, vaswani2017attention,vaswani-etal-2018-tensor2tensor}, online translation systems can provide users with fast and accurate translation results, making them widely used. However, the potential privacy concerns arising from the service mode of online translation systems are also receiving increasing attention. As shown in Fig.~\ref{translation_paradigm}, when using online translation systems, users typically need to upload the complete text over the network to the translation models located on servers, which raises concerns about privacy leakage for users who value data privacy. Although translation service providers make privacy protection claims, in some scenarios where data privacy is highly valued, data is not allowed to be transmitted to external environments due to the potentially serious consequences it may entail~\footnote{For example, there was an incident at the renowned Norwegian oil company Statoil where a significant amount of private information was leaked due to the use of online translation tools. This has led some financial institutions to prohibit their employees from using external online translation tools. This case shows that potential privacy leakage objectively hinders the use of online translation systems in privacy-conscious scenarios. You can obtain more details from \url{https://www.csoonline.com/article/563519/data-breached-in- translation.html}}.

To alleviate this predicament, online translation services need to introduce privacy protection mechanisms targeting translation model inference. Currently, although privacy protection during the inference stage of NLP models has recently received widespread attention, some methods based on sanitization of private information~\cite{huang-etal-2023-privacy,chen2023hide,wang2023decodingtrust}, secure multiparty computation~\cite{adams-etal-2021-private,NEURIPS2022_64e2449d,cryptoeprint:2023/1269}, and local differential privacy~\cite{igamberdiev-habernal-2023-dp,feyisetan2020privacy} have been proposed. However, they are mainly applied to tasks such as text classification, text summarization, question answering, and inference. Machine translation still lacks relevant research work in this regard, including a clearly defined privacy-preserving task for the inference of machine translation model, dedicated datasets, corresponding evaluation metrics and baseline methods. The absence of these key elements has greatly hindered research on privacy protection in the inference process of translation models in the field of machine translation.

\begin{CJK*}{UTF8}{gbsn}

\end{CJK*}

To bridge these gaps, we formally define a novel machine translation task, Privacy-Preserving Machine Translation (PPMT), aiming to protect the private information in text during the machine translation model inference stage. To support the PPMT task, we constructed dedicated datasets, designed evaluation metrics for privacy leakage, and provided some sanitization-based methods as baseline methods. 

Among these elements, the most challenging aspect is constructing a specialized dataset. For the PPMT task, we need to annotate the privacy content in bilingual corpora. However, privacy has a complex and diverse definition. While the GDPR enacted by the European Union provides some definitions of privacy content, these definitions primarily revolve around personally identifiable information, while the privacy that online translation systems need to protect is more focused on trade secrets. Different users have varying definitions of trade secrets, making it difficult to comprehensively and accurately define which content in the text belongs to privacy content. To reduce the difficulty of dataset construction, we consider that named entities constitute a significant portion of business information and personally identifiable information. Therefore, we only annotate the named entities in the text for privacy labeling, filtering out the named entities that do not belong to privacy. On the other hand, since there are differences in defining privacy content between general and vertical domains, to avoid a single-domain dataset, we annotate data from both general and financial domains. Ultimately, we manually annotate three bilingual datasets as the test datasets for this task. 

In addition to the dataset, we also provide evaluation metrics specific to this task. When designing the metrics, we consider two aspects: one is to measure the effectiveness of the privacy protection method, and the other is the translation performance of the translation model after applying the privacy protection method. Relevant work has shown that the introduction of privacy protection mechanisms can reduce the model's performance. A good privacy protection method should minimize the adverse impact on the model's performance while protecting privacy.

For the baseline methods, we were inspired by the sanitization of private information in other works~\cite{huang-etal-2023-privacy,chen2023hide,wang2023decodingtrust}, and designed a framework based on privacy sanitization, using different privacy sanitization strategies as different baseline methods. Specifically, this framework introduces a privacy sanitization module at the machine translation model inference stage to sanitize the private information in the text to be translated. Then the sanitized text is input into the translation model that has been adapted to take sanitized text as input. The privacy information is translated separately by a privacy translation module that the translation model cannot access, ensuring that the privacy information is not leaked to the translation model. Then a merge module combines the translation results returned by the translation model and the privacy translation to form the final translation result. In the specific implementation, we use the classic Transformer~\cite{vaswani2017attention} as the backbone translation model, and propose some model enhancement strategies during the training stage to adapt to this framework.

In summary, our main contributions are as follows:
\begin{enumerate}
    \item Given the lack of research on privacy protection during the inference stage of translation models in the field of machine translation, we first propose Privacy-Preserving Machine Translation (PPMT), a new machine translation task that aims to preserve privacy during the inference stage of the machine translation models.
    \item We annotate three bilingual datasets as test datasets for this task, using named entities as candidate privacy content in the text. The data includes both general domain and financial domain data, addressing the issue of the lack of specialized datasets in this field.
    \item Considering the lack of dedicated evaluation metrics, we provide evaluation metrics to measure the ability of privacy-preserving machine translation methods. These metrics aim to capture both the effectiveness of the privacy protection mechanism and the translation performance of the model after the privacy protection is applied.
    \item We provide a privacy sanitization-based protection framework, and propose some sanitization strategies as specific baseline methods. We use the Transformer as the backbone translation model, and obtain the performance of these strategies based on the measurement metrics we proposed as baseline scores.
\end{enumerate}
\section{Related Work}

\subsection{Privacy-Preserving for General NLP Tasks}
With the continuous development of AI technology, human production, and life have obtained great assistance from AI technology. However, many new legal and ethical issues~\cite{amodei2016concrete, floridi2019establishing, arrieta2020explainable,carlini2021extractingtrainingdatalarge} have followed, and one of the most widely concerned issues is the privacy preservation of AI models. At present, many research works have noticed this problem~\cite{Coavoux2018PrivacypreservingNR, elmahdy-etal-2022-privacy, ponomareva-etal-2022-training, zhan2023measuring}, and have proposed several solutions (such as federated learning, differential privacy~\cite{sousa2022keep}) for their focused areas. In detail, ~\cite{jedrzejewskiprivacy} systematically discuss the challenges and solutions in the research field of Privacy-Preserving Natural Language Processing.~\cite{comi2020herbert} introduce a privacy-preserving natural language processing solution called HErBERT, capable of performing text classification on encrypted data. The proposed solution relies on Homomorphic Encryption to process the encrypted data.~\cite{qu2021natural} study the impact of applying differential privacy to BERT fine-tuning in NLU applications and further proposed privacy-adaptive LM pretraining methods. This achieved significant improvements in BERT's performance while protecting privacy, and also quantified the level of privacy preservation.~\cite{yin-habernal-2022-privacy} discuss to what extent the privacy of pre-training data can be guaranteed while simultaneously improving downstream performance on legal tasks without the need for additional labeled data.~\cite{nagy2023privacy} propose a privacy-preserving federated learning framework that combines bitwise quantization, local differential privacy, and feature hashing techniques, capable of handling any type of input features. This work emphasizes the usability of the framework with natural language data and evaluates it on a sentiment analysis task.

\subsection{Privacy-Preserving for Machine Translation}
While there are some works~\cite{shokri2015privacy,sadat2019privacy,comi2020herbert,habernal2023privacy,nagy2023privacy,sousa2023keep} focusing on privacy protection for natural language processing models, they primarily concentrate on protecting the training data, and there is limited research specifically addressing machine translation. In the field of machine translation, although neural machine translation has achieved amazing performance in recent years and has been widely used in human production and life, the research on privacy-preserving machine translation is still in its infancy, resulting in the relatively scarce relevant literature. 

~\cite{kamocki-oregan-2016-privacy} indicated that free online translation systems may suffer legal issues such as the processing of personal data and checked how the free online machine translation systems fit in the European data protection framework from both the user and the service provider's views.~\cite{9099258} proposed a natural language processing system based on multi-party privacy-preserving to avoid privacy implications in the collection of large-scale data used to train deep learning models. Although this work tests the system on machine translation tasks, it only demonstrates the system's performance on runtime and message communication without showing the translation performance. To decrease the legal risks in collecting training data from multiple institutes,~\cite{10.1007/978-3-030-90888-1_18} proposed a federated neural machine translation model, which could train a translation model without sharing raw data from participants. After that,~\cite{He2022SyntheticPT} utilized synthetic data to train the pre-train models to alleviate such concerns.

The above works mainly focus on privacy-preserving in training data. However, as an important scenario in which commercial translation systems provide services in reality, translating a sentence without seeing its private information still lacks corresponding attention. In this paper, we formally define this task and propose a complete benchmark from data, and metrics to baseline method to bridge this gap. 

\section{Benchmarking PPMT}
In this section, we will introduce privacy for machine translation task, the formalization and goal of \textbf{P}rivacy-\textbf{P}reserving \textbf{M}achine \textbf{T}ranslation (\textbf{PPMT}) task, and evaluation metrics we use for this task.

\subsection{Privacy Content for Machine Translation Task}
Existing privacy protection measures often focus on protecting personally identifiable information (such as names, ID cards, home addresses, phone numbers, etc.) that can identify individuals. However, for online translation systems, the concept of privacy also includes information that individuals or organizations do not want to disclose or share with others. In the context of online translation systems, privacy can involve trade secrets, sensitive business information, or other confidential internal information of organizations. Sensitive information in the text typically consists of a set of terms that collectively describe a particular idea, rather than individual terms~\cite{garcia2017identifying}. As shown in Fig.~\ref{translation_paradigm}, the words "Apple" and "Friday" together describe a significant business action that Apple might undertake, and the conveyed information is private to Apple. However, \textbf{if either "Apple" or "Friday" is removed from the sentence, the sensitivity of the sentence significantly decreases.} In this sense, "Apple" and "Monday" form a set of terms that describe the sensitive information in the sentence. This set is the private content that needs protection for the translation system.

Although we defined \textbf{a set of terms that collectively describe the sensitive information contained in the sentence as the privacy content or privacy-related terms} that online translation systems need to protect, in practice, it has been challenging to establish a comprehensive and consistent definition of this term set due to different interpretations by individuals. To concretize the privacy in machine translation, based on the understanding of privacy content mentioned above, we conducted a manual investigation of hundreds of sentences and found that the majority of privacy content appears in the form of specific types of named entities, such as person names, locations, organizations, quantities, and time. This observation aligns with the findings of~\cite{9152761}.~\footnote{There is a discussion between privacy information and entities in \url{https://towardsdatascience.com/remove-personal-information-from-a-text-with-python-part-ii-ner-2e6529d409a6}.} Table~\ref{privacy cases} shows some examples of texts including privacy content.

Taking into account these factors, in order to provide a clear, specific, and feasible annotation protocol for dataset construction, we have limited the privacy content for translation system to the scope of named entities including \textbf{person names, locations, organizations, quantities, and time}. In the annotation process, we mainly select entities belonging to the privacy content from these named entities. However, we must acknowledge that this specification restricts the scope of privacy content. It is a compromise we have made due to the current limitations of resources. In future work, we will further expand the scope of privacy content.

\begin{table*}[!htb]
  \centering
  \resizebox{\linewidth}{!}{
  \begin{tabular}{ccc}
  \toprule
   \multicolumn{2}{c}{Example 1} &  \textcolor{red}{The United Nations International Atomic Energy Agency} ( \textcolor{red}{IAEA}) released a report on the \textcolor{red}{26th} stating that \textcolor{red}{Iran} \\ & &  has started to use advanced centrifuges to accelerate the production of enriched uranium , further violating the nuclear agreement reached with world powers in \textcolor{red}{2015} . \\
   \midrule
   \multicolumn{2}{c}{Example 2} & \textcolor{red}{Warren Buffett} to sell \textcolor{red}{one million} shares of \textcolor{red}{Google} Inc next \textcolor{red}{Monday} . \\
   \midrule
   \multicolumn{2}{c}{Example 3} & \textcolor{red}{The Hong Kong Monetary Authority} will introduce a new securities trading policy on \textcolor{red}{1 April} and will hold a press conference at \textcolor{red}{3 pm} on the same day to explain the new policy . \\
  \bottomrule
  \end{tabular}
  }
  \caption{Cases of texts including privacy information. The red parts in listed sentences are privacy-related terms in our definition.}
  \label{privacy cases}
\end{table*}

\subsection{Task Formalization}
Given a bilingual sentence pair $(X, Y)$, where $X = (x_1, x_2, ..., x_M)$ is the source language  sentence, $Y = (y_1, y_2, ..., y_N)$ is the target language sentence, the general translation process without considering privacy information could be represented as:
$$P(Y) = \prod_{t=1}^N p(y_t|X, Y_{<t})$$, where $Y_{<t}$ means prefix tokens in $Y$ before timestep $t$.

In the privacy preservation situation, both the source and target language sentences contain privacy-related terms, which usually are in the form of phrases. Here, we denote the privacy information in the source sentence as $X_p = [(x_i x_{i+1} x_{i+2} ...), (x_j x_{j+1} x_{j+2} ...), ...]$ and their corresponding translations in the target sentence as $Y_p = [(y_i y_{i+1} y_{i+2} ...), (y_j y_{j+1} y_{j+2} ...), ...]$. In the PPMT task, we expect to obtain good translations and reduce the disclosure of these privacy-related terms in the inference process of the translation system. Specifically, in the online translation scenario, we want to have as few privacy-related terms as possible appearing in the transmission of texts and in the reception of translations. Taking the sentence in Fig.~\ref{translation_paradigm} as an example, the user is using an online translation service to send in a sentence containing the privacy-related terms 
\begin{CJK}{UTF8}{gbsn} "苹果 (Apple)" and "周五 (Friday)"\end{CJK}, and PPMT needs to ensure that, as far as possible, these two terms and their corresponding translations "Apple" and "Monday" do not appear in the external environment (cloud server), while at the same time obtaining a high-quality translation of the sentence.

\subsection{Evaluation Metrics}
\begin{algorithm} 
	\caption{Calculation of Privacy Exposure Rate (PER)} 
	\label{alg3} 
	\begin{algorithmic}
        \STATE 1. Count the number of tokens in $X_p$ that appear in the text sent to the translation model, $N_{X_e}$ and the number of tokens in $X_p$, $N_{X_p}$
        \STATE 2. Count the number of tokens in $Y_p$ that appear in the text generated by translation model, $N_{Y_e}$ and the number of tokens in $Y_p$, $N_{Y_p}$
        \STATE 3. Count the ratio of privacy tokens in the source text transmission process to all privacy words in the source text: $R_{X} = \frac{N_{X_e}}{N_{X_p}}$
        \STATE 4. Calculate the ratio of privacy tokens in the translation returning process to all privacy words in the reference sentence: $R_{Y} = \frac{N_{Y_e}}{N_{Y_p}}$
        \STATE 5. Calculate the final Privacy Exposure Rate $PER = \frac{(R_{X} + R_{Y})}{2}$ as the final metric.
	\end{algorithmic} 
\end{algorithm}
The two core requirements of this task are to prevent the disclosure of private information and to obtain a high-quality translation. Therefore, we can evaluate the method from two perspectives, the degree of protection of private information and the quality of the final translation.

For the evaluation of the quality of the final translation, we utilize general metrics in machine translation tasks, including the SacreBLEU~\cite{post-2018-call}, chrF~\cite{popovic-2015-chrf} and BLEU~\cite{papineni-etal-2002-bleu}.

For evaluation of the degree of exposure of privacy information, we consider two kinds of privacy breaches: one occurs during the transfer of the source language text from the user to the server and one occurs when the server returns the translated text. Here, we use a comprehensive metric, the Privacy Exposure Rate (PER), to measure the extent of privacy breaches throughout the process. The idea behind this metric is that the more privacy-related tokens appear in the text uploaded by the user and the text returned by the translation system, the more serious the privacy breach. The PER calculation processes are performed in Algorithm~\ref{alg3}. The smaller the PER, the higher the level of privacy protection. In addition, when actually calculating PER, we ignore tokens that are marked as privacy-related but have no real meaning, such as "the", "of", and, "etc".

\section{Data Annotation}
\begin{table*}[htb]
    \centering
    \scalebox{0.8}{
    \begin{tabular}{cccccccc}
        \toprule
            & Language & Total Sentences & Person Name & Location & Organization & Time & Quantity  \\
        \midrule
        \multirow{2}{*}{WMT18} & Zh & 1928 & 2053 & 1551 & 1333 & 1347 & 1307 \\  & En & 1928 & 1960 & 1516 & 1306 & 1211 & 1188 \\

        \multirow{2}{*}{WMT20} & Zh & 1964 & 1088 & 1813 &  1230 & 1372 & 1261 \\  & En & 1964 &  1068 & 1751 & 1215 & 1245 & 1139 \\
 
        \multirow{2}{*}{Finance} & Zh & 1744 & 218 & 316 & 1949 & 1405 & 988 \\  & En & 1744 & 197 & 300 & 1889 & 1342 & 930\\
        \bottomrule
    \end{tabular}}
    \caption{The statistics of human-annotated test datasets, including the number of total sentences and the number of privacy entities with different entity types.}
    \label{dataset}
\end{table*}

\subsection{Overview}
To provide the dedicated datasets for the PPMT task, we hire professional human annotators to annotate the privacy-related terms in the parallel corpus to obtain test datasets. In the following parts, we will introduce details for data collection and human annotation and list the statistical information of annotated datasets. All datasets will be made public after the review.

\subsection{Data Collection}
Before manually annotating the parallel corpus, we need to collect the original parallel corpus. Considering that different domains have different distributions of privacy information, we collect parallel corpus from general and vertical areas for a comprehensive evaluation. For the general domain, we choose the test datasets of WMT18 Zh-En~\cite{bojar-etal-2018-findings} and WMT20~\cite{barrault-etal-2020-findings} Zh-En as the original datasets. For the vertical domain, we choose the texts from the financial domain and collect publicly available financial reporting texts from the web in both English and Chinese.

\subsection{Annotation Process}
Human annotators are required to annotate privacy-related terms in each sentence pair if the sentence pair contains privacy. Each sentence pair is handled by 3-5 annotators. The complete annotation process includes 3 steps. \textbf{Firstly}, human annotators need to determine whether the given sentence pair involves privacy. We consider a sentence to contain privacy if more than half of the annotators consider it to contain privacy. \textbf{Secondly}, the sentence pair containing privacy will then be processed by Texsmart\cite{texsmart2021} to identify the named entity in that sentence pair. This step is intended to give different annotators the same set of privacy-related term candidates, preventing confusion in their annotation due to inconsistent understanding of the scope of the entity. \textbf{Thirdly}, human annotators need to select privacy-related terms from recognized named entities. In this process, if a privacy-related term is accepted by more than two annotators, it would be regarded as a privacy-related term.

Also, sentence pairs with two non-matching sentences are discarded and every entity that meets our definition of privacy is labeled. A complete privacy-related term annotation includes the privacy \textbf{term's string, starting position, string length, and entity type}.

\subsection{Statistics}
The dataset statistics are in Table~\ref{dataset}. According to this table, we could find that the three datasets have different distributions over the types of privacy information. The majority of entity types in WMT18 are "Person Name" and "Location". In WMT20, the main components are "Place" and other entity types have the nearly same number of privacy terms. For the Finance dataset, the "Organization", "Time" and "Quantity" is the most frequent entity types.
\begin{CJK*}{UTF8}{gbsn}
\begin{figure*}
    \centering
    \includegraphics[width=\linewidth]{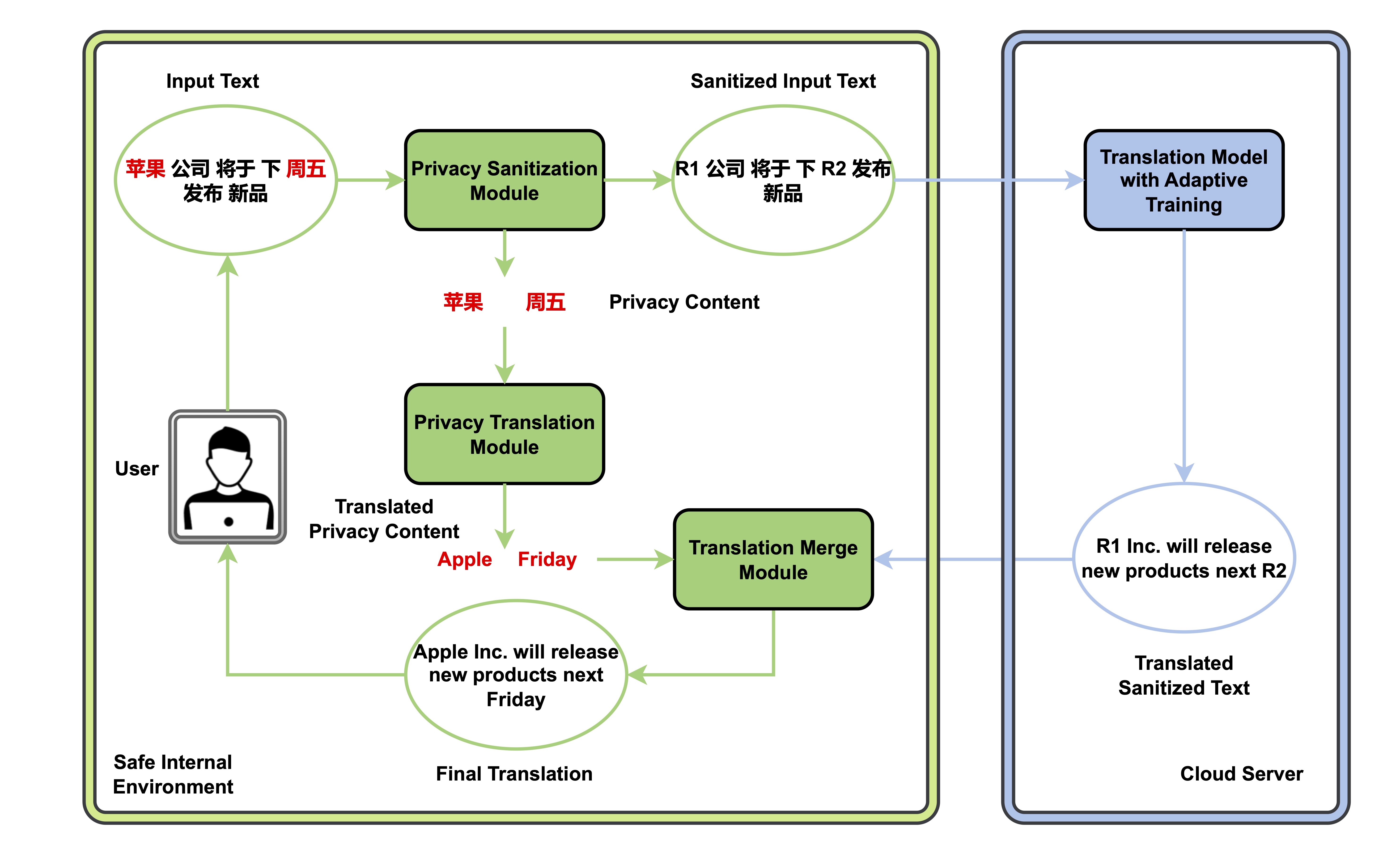}
    \caption{This is our proposed framework. In this work, we use the client and cloud server to indicate safe internal and risky external environments. The sanitized sentence is sent to the server from the client and the client receive the translation of sanitized text. R1 and R2 represent two different placeholders of two privacy terms "苹果" and "周五" assigned by privacy sanitization module.}
    \label{framework}
\end{figure*}

\section{Benchmark Models}
\subsection{Overview}
To address the lack of baseline methods for PPMT, we design a privacy sanitization-based framework and propose several sanitization strategies to serve as specific baseline methods. These methods are straightforward and easy for subsequent researchers to implement.

In detail, to reduce the degree of privacy exposure, a straightforward approach is to scoop out the privacy-related terms from the text and send the non-private parts to the external translation system. This approach can significantly reduce the extent of privacy breaches, but it can also be detrimental to the quality of the translation due to the loss of private information and the disruption of the context of non-privacy information. It would be better if we could translate the privacy-related terms of texts to be translated in a way that reduces the extent of privacy leakage, then obtain a high-quality translation of the non-private parts, and finally stitch the two together. Based on this idea, we design a privacy sanitization framework, which utilizes the sanitization method to process the input text to prevent external translation model from obtaining the privacy information, sends the sanitized text to the translation model on the server for good translation, obtains the translation of privacy information with a privacy translation module and combine two kinds of translations with a translation merge module. 

Here, we take the online translation service scenario as an example to describe the detailed workflow of the framework. As shown in Fig.~\ref{framework}, this framework consists of four components: Privacy Sanitization Module, Privacy Translation Module, Translation Merge Module and Translation Model with Adaptive Training. The first three modules are located in a secure internal environment, such as the user client. The last module is located on the cloud server of the translation service provider, and we do not want private content to be accessed by this module.

After receiving the user's input text, privacy sanitization module replace two privacy terms "苹果" and "周五" with two placeholders "R1" and "R2". The privacy terms "苹果" and "周五" are translated as "Apple" and "Friday" by the privacy translation module. The sanitized input text is sent to a translation model that adapts to this input, which has undergone adaptive training to better translate the sanitized text. Finally, the translations of privacy content and sanitized text are merged by the translation merge module in the safe internal environment to produce the final translation for the user. Throughout the entire process, private content and its translation are always in a secure environment, so they can be well protected.

We have outlined the modules included in the framework and the entire workflow in the previous chapter. In the following, we will provide a detailed introduction to the mechanisms and implementations of each module.

Based on the above description of how our framework works, it is easy to see that the biggest challenge of the framework is: \textbf{How to train a translation model to produce high-quality translations for non-privacy information}, which includes two aspects: on the one hand, the translation of the content of the non-private parts should be accurate, and on the other hand, the placeholders are assigned to the appropriate positions.
We will describe in detail how we trained such a translation model in the next section.

Compared to other methods such as secure multiparty computation and local differential privacy, this framework does not require any modifications to the architecture of the translation model itself, nor does it require the user side to bear significant computational costs.

\subsection{Privacy Sanitization Module}
The role of the privacy sanitization module is to process the private content in the input text to ensure that the translation model cannot access this private content. Specifically, this module replaces the private content in the text with placeholders, and then inputs the replaced text into the translation model. In this article, we propose three replacement strategies with different placeholders:

\subsubsection{Entity-based Placeholder}

As shown by Fig~\ref{framework}, the translation model takes non-privacy information connected by placeholders as input. In this case, translation models trained on a standard bilingual corpus are generally unable to translate texts including placeholders well because the training corpus has not seen these placeholders. However, if we use the special entities as placeholders and replace the privacy information in the sentence with entities having the same entity type, we can still use the translation model trained on standard bilingual data to deal with these replaced sentences. After obtaining this translation model's generation, we could replace these entities' translations in the generation with the original privacy-related terms' translations in the client to produce the final translation.

\subsubsection{Dictionary-based Placeholder}  

The entity placeholder replacement method is simple to implement. However, as the same entity may correspond to multiple translation results, this makes it more difficult to merge the translations of privacy and non-privacy information. This problem can be alleviated if the translation model can produce a relatively fixed output for certain entities. To this end, we use the named entity recognition tool~\cite{texsmart2021} and word alignment tool~\cite{chen-etal-2021-mask} to collect some high-frequency entity translation pairs (such as \textbraceleft 联合国 (the United Nations), the United Nations\textbraceright) from the standard training corpus to create a translation dictionary for each type of entity. In the process of privacy sanitization, we randomly select the same type entity to replace the privacy entity in the input text. A replacement case is listed in Fig~\ref{dict}.

\subsubsection{Tag-based Placeholder}
\begin{figure}
    \centering
    \includegraphics[width=\linewidth]{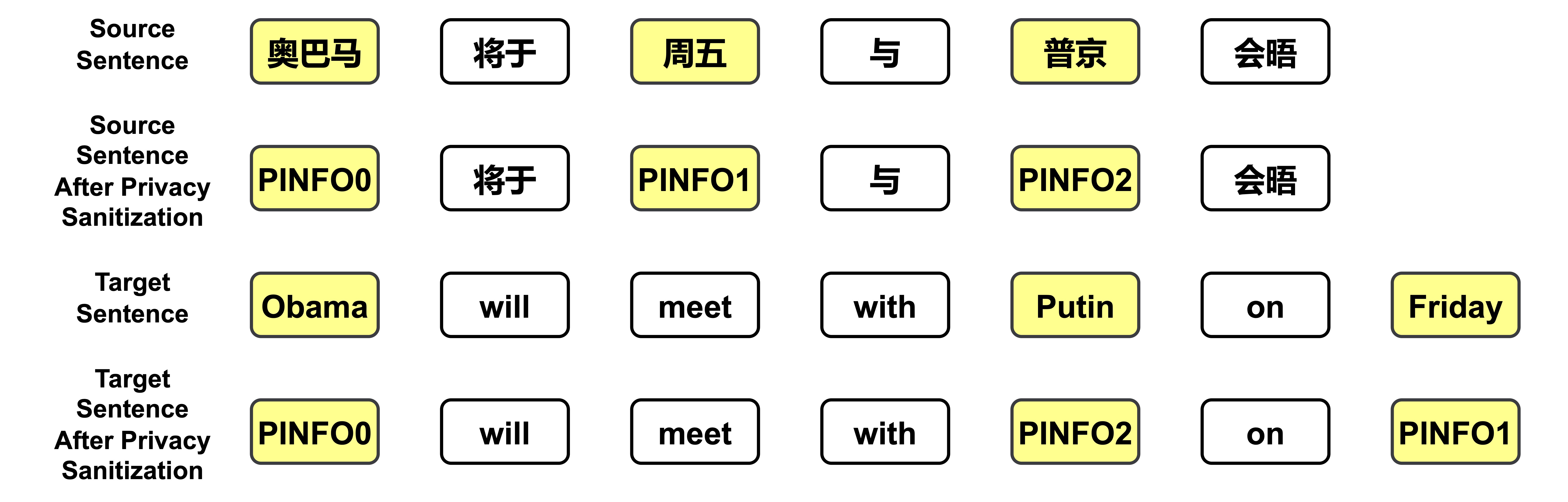}
    \caption{A case of replacement with the tag-based placeholder. The privacy parts are in yellow.}
    \label{tag}
\end{figure}

Constructing a large translation dictionary from a large number of parallel corpora can sometimes be time-consuming. Therefore, we propose a tag based replacement method, which replaces the privacy content in the input text with a fixed tag so we do not have to construct a high-frequency translation dictionary for each type of entity. Compared with the dictionary-based placeholder, this kind of placeholder is more general. In this paper, we use the tag "PINFO" as the placeholder, and a replacement case is shown in Fig~\ref{tag}.

\subsubsection{Comparison between Placeholders}

The entity-based placeholder is randomly selected from entities of the same type as the one being replaced, whereas the tag-based placeholder uses a fixed symbol [PINFO] followed by a sequential number. Compared to the tag-based method, sentences generated with entity-based placeholder replacement more closely resemble natural language and do not require generating specialized training data to retrain the translation model.

The key difference between dictionary-based and entity-based placeholders lies in the source of replacement: dictionary-based placeholders are selected from a relatively fixed and small-sized dictionary, whereas entity-based placeholders are randomly chosen from a much larger pool of entities. As a result, sentences with dictionary-based replacements tend to frequently include certain recurring entities, leading to a skewed textual distribution.

Therefore, we performed an adaptive training for dictionary/tag-based placeholder methods but did not apply it to the entity-based approach, as the latter already generates texts that are more natural and distributionally consistent.

\begin{figure}
    \centering    \includegraphics[width=\linewidth]{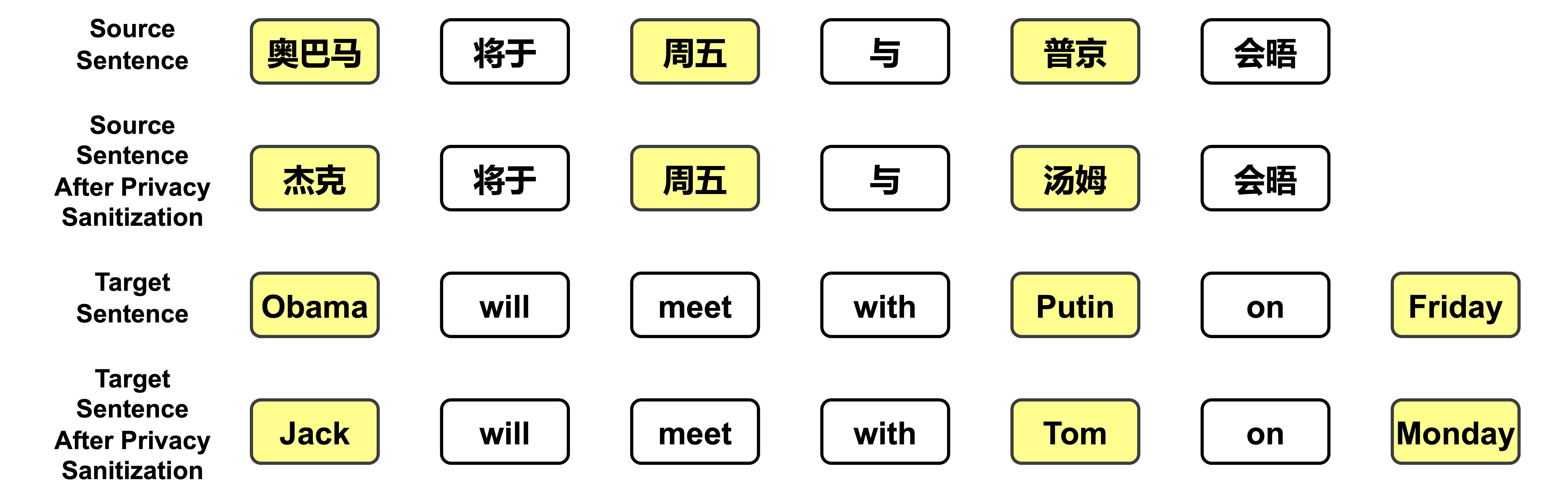}
    \caption{A case of replacement with the dictionary-based placeholder. The privacy parts are in yellow.}
    \label{dict}
\end{figure}

\begin{figure}
    \centering
    \includegraphics[width=\linewidth]{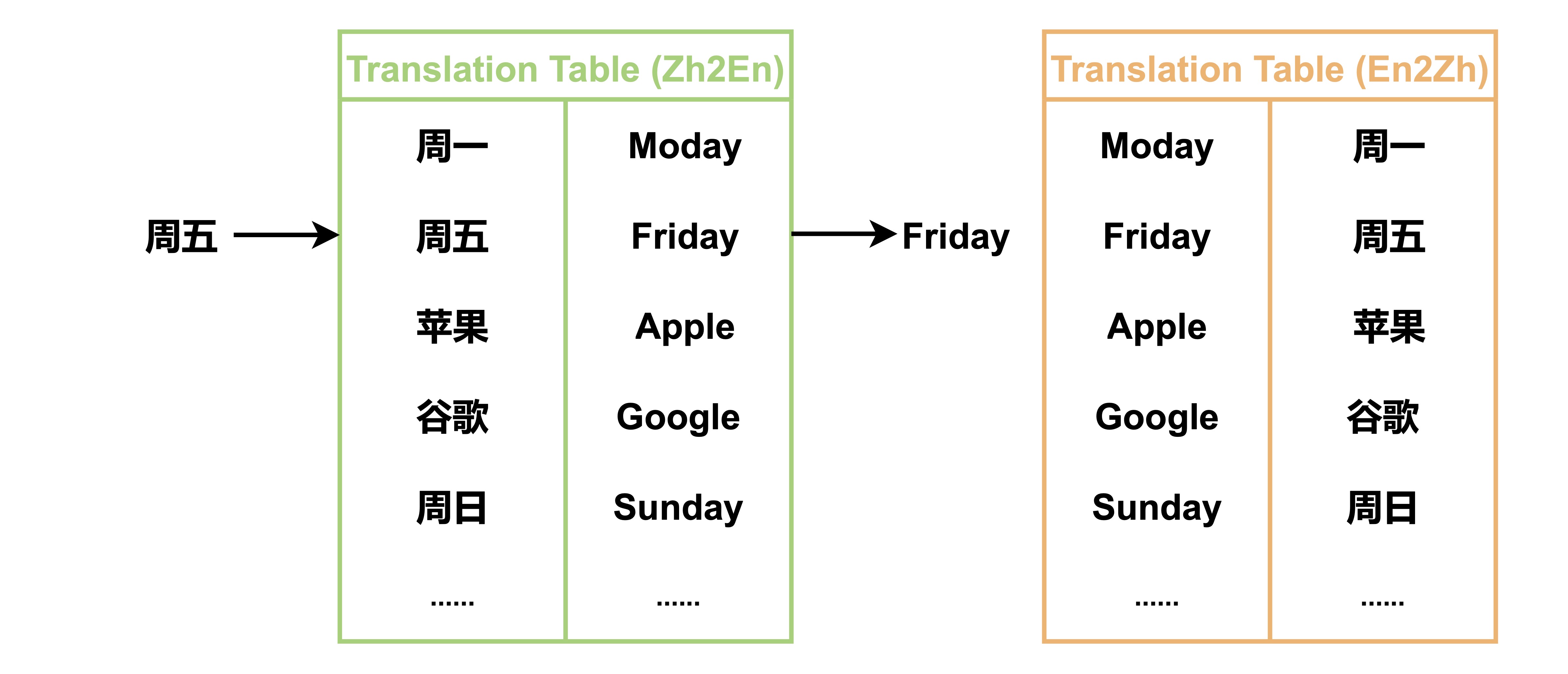}
    \caption{A case of how to obtain the privacy information's translation according to a phrase translation table.}
    \label{look_up_table}
\end{figure}

\subsection{Privacy Translation Module}
For the translation of privacy information, given that the user side cannot afford the high storage and computational overhead and the majority of privacy terms are phrases with relatively fixed translations, the privacy information is translated by looking up a phrase translation table (A case is shown in Fig~\ref{look_up_table}) in the client thereby mitigating privacy leakage. Here, we use the translation dictionary in dictionary-based placeholder method to serve as the phrase translation table.

\subsection{Training of Translation Model}
\subsubsection{Adaptive Training for Translation Model}
We have found through experiments that the translation quality of translation model trained on the standard bilingual corpus is poor due to the inconsistent distribution of privacy terms between the training data and real input. This problem could be alleviated by constructing new training data more similar to the real input texts. In detail, we replace the named entities of the bilingual training data with translation pairs from these dictionaries (as the case shown in Fig~\ref{dict}) or tags (as the case shown in Fig~\ref{tag}) to construct a new training dataset. This new training dataset is utilized to train a translation model. With this translation model, we could obtain a high-quality translation of the input received by the server and could merge this translation and the translation of privacy information more easily. We call this trick \textbf{adaptive training}.

\subsubsection{Data Augmentation for Translation Model with Adaptation Training}
In our experiments, we found that while the translation with model adaptation training can handle replaced sentences well, it is somewhat less effective than the standard training model when faced with sentences that have only a small amount of privacy or even no privacy to replace. To further improve the translation quality of our translation model, we apply a data augmentation technique to the training process of the translation model with adaptation training. The idea is to add adversarial data. Specifically, we mix sentence pairs in newly constructed training datasets and corresponding original sentence pairs together and finally use these mixed training datasets to train the translation model. In this way, the trained translation model could translate sentences with a small amount of privacy information well and thus achieve better translation performance.

\subsection{Merge Module}

The translations of the non-private parts are returned to the user client and combined with the translation of the privacy information to form a complete sentence translation. To obtain a precise combination, we replace privacy-related terms in input sentences with placeholders ("苹果" and "周五" are replaced by R1 and R2) and the translation model after adaptive training can retain the placeholders in the input at the appropriate positions in the output. In this case, to obtain the final translations, we just need to replace the placeholders in the returned non-private translations with the translations of corresponding privacy-related terms in the client. If the translation model is trained on standard bilingual corpus and the privacy terms in the input text are replaced by several randomly selected entities, we need to replace translation of these selected entities with the translation of privacy terms.

\end{CJK*}
\section{Experiment}
Here, we evaluate the translation and privacy protection performance of several baselines corresponding to the proposed framework, comparing their computational efficiency with classical methods such as local differential privacy, secure multiparty computation, and large model-based local deployment strategies. Subsequently, we conduct a series of analytical experiments, including: verifying the necessity of adaptive training and data augmentation, analyzing the impact of different Named Entity Recognition (NER) privacy identification methods, and examining the influence of NER errors. Furthermore, we expand the scope of privacy by adding more entity types and contextual privacy, and thus label two new datasets. Then, we validate the effectiveness of the framework in broader privacy scenarios. Finally, we conduct a series of case studies to analyze the advantages and limitations of our framework.

\subsection{Datasets for Adaptation Training}

As mentioned in the benchmark models section, we need to construct new training datasets for the translation model's adaptation training. For this purpose, we use Texsmart~\cite{texsmart2021} to recognize the entities named in the training data of WMT18, WMT20 and financial datasets, and then train a word alignment tool (Mask-Align~\cite{chen-etal-2021-mask}) to obtain alignment relations between source language entities and target language entities. Then, we replace the named entities with placeholders (dictionary-based and tag-based) to construct new training datasets for adaptation training. In addition, the training data of the financial domain are not sufficient (only 50k) to train a large translation model. Therefore, we use the financial training data to fine-tune a translation model trained on the large-scale corpus. In our experiments, we fine-tune the translation model trained on the WMT18 data.

\subsection{Backbone Translation Models}
Our framework performance experiments compare the performance of three translation systems with different training approaches. The first is a translation model trained on the standard bilingual parallel corpus, which accepts input as normal sentences without any replacement. The second translation system takes test sentences after replacing privacy information with dictionary-based placeholders as inputs, but uses the adaptive training and data augmentation methods mentioned above. The third translation system takes as input sentences after replacing the privacy-related terms with tag-based placeholders, and also utilizes adaptation training and data augmentation methods. Each of these three translation systems corresponds to the three translation models. We use \textbf{Trans, Tag-Trans, and Dict-Trans} to represent these three translation models.

\subsection{Training Settings}
\label{training_settting}
We choose the big version of the standard transformer as the translation model, which includes 6 encoder layers and 6 decoder layers. Each attention layer has 16 attention heads and a hidden size of 1024. All models in this paper are implemented with THUMT~\cite{tan-etal-2020-thumt}. We use the shared BPE~\cite{sennrich2016neuralmachinetranslationrare} (implementation is based on subword-nmt) vocabulary with a vocabulary size of 32K and translation models are trained for 400,000 steps on 4 V100s with a batch size of 4096 on WMT18 and WMT20 datasets. The optimizer we use is Adam~\cite{kingma2017adammethodstochasticoptimization} and learning rate is set to 0.0001. We fine-tune the translation models trained on the WMT18 dataset with financial data for 100,000 steps. Other settings in fine-tuning are consistent with the training on the WMT18 dataset.

\subsection{Test Scenarios}

\begin{table*}[!htb]
  \centering
  \resizebox{\linewidth}{!}{
  \begin{tabular}{cccccccccccccccccccccccccccc}
  \toprule
   \multicolumn{2}{c}{\multirow{3}*{\shortstack{Placeholder}}} & \multicolumn{2}{c}{\multirow{3}*{\shortstack{Translation Model}}} & \multicolumn{8}{c}{WMT18}  & \multicolumn{8}{c}{WMT20}  & \multicolumn{8}{c}{Finance} \\
   
   \cmidrule(lr){5-12}\cmidrule(lr){13-20}\cmidrule(lr){21-28}
   \multicolumn{4}{c}{} & \multicolumn{4}{c}{\ZhEn} & \multicolumn{4}{c}{\EnZh}  & \multicolumn{4}{c}{\ZhEn} & \multicolumn{4}{c}{\EnZh}  & \multicolumn{4}{c}{\ZhEn} & \multicolumn{4}{c}{\EnZh} \\
   \cmidrule(lr){5-8}\cmidrule(lr){9-12} \cmidrule(lr){13-16} \cmidrule(lr){17-20} \cmidrule(lr){21-24}\cmidrule(lr){25-28}
   \multicolumn{4}{c}{} & SacreBLEU  & chrF & BLEU  & PER & SacreBLEU  & chrF & BLEU & PER & SacreBLEU  & chrF & BLEU & PER & SacreBLEU  & chrF & BLEU & PER & SacreBLEU  & chrF & BLEU & PER & SacreBLEU  & chrF & BLEU & PER \\
   
   \midrule
   
   \multicolumn{2}{c}{-} & \multicolumn{2}{c}{Trans} & 22.80 & 51.40 & 22.61 & - & 18.60 & 29.50 & 18.18 & - & 27.00 & 55.50 & 27.90 & - & 19.20 & 33.10 & 19.06 & - & 47.40 & 71.20 & 48.19 & - & 39.60 & 47.30 & 38.57 & - \\
   \midrule
   \multicolumn{2}{c}{Entity} & \multicolumn{2}{c}{Trans (3 types)} & 19.50 & 44.10 & 19.33 & 0 & 15.80 & 25.70 & 15.62 & 0 & 23.90 & 49.30 & 24.72 & 0 & 16.80 & 27.30 & 16.69 & 0 & 44.70 & 67.40 & 46.57 & 0 & 37.00 & 44.50 & 36.82 & 0  \\
   \midrule
   \multicolumn{2}{c}{Entity} & \multicolumn{2}{c}{Trans (5 types)} & 18.90 & 42.80 & 18.76 & 0 & 14.70 & 23.90 & 14.54 & 0 & 23.40 & 48.30 & 24.13 & 0 & 16.20 & 26.20 & 16.01 & 0 & 43.80 & 65.90 & 45.53 & 0 & 35.70 & 42.90 & 35.46 & 0 \\
   \midrule
   \multicolumn{2}{c}{Dictionary} & \multicolumn{2}{c}{Dict-Trans (3 types)} &
   21.40 & 48.40 & 21.20 & 0 & 17.30 & 27.90 & 17.08 & 0 & 25.70 & 52.90 & 26.50 & 0 & 18.90 & 33.10 & 18.77 & 0 & 47.00 & 70.70 & 48.87 & 0 & 38.90 & 46.80 & 38.78 & 0 \\
   \midrule
   \multicolumn{2}{c}{Dictionary} & \multicolumn{2}{c}{Dict-Trans (5 types)} &
   20.90 & 47.60 & 20.81 & 0 & 16.60 & 26.90 & 16.45 & 0 & 25.30 & 52.00 & 26.01 & 0 & 18.20 & 31.80 & 17.98 & 0 & 45.80 & 69.00 & 47.69 & 0 & 37.90 & 45.70 & 37.63 & 0 \\
   \midrule
   \multicolumn{2}{c}{Tag} & \multicolumn{2}{c}{Tag-Trans (3 types)} & 21.40 & 48.50 & 21.24 & 0 & 17.40 & 28.10 & 17.17 & 0 & 25.60 & 52.50 & 26.34 & 0 & 18.80 & 32.90 & 18.67 & 0 & 46.80 & 70.50 & 48.69 & 0 & 38.90 & 46.60 & 38.65 & 0 \\
   \midrule
   \multicolumn{2}{c}{Tag} & \multicolumn{2}{c}{Tag-Trans (5 types)} & 20.80 & 47.20 & 20.62 & 0 & 16.70 & 27.10 & 16.58 & 0 & 25.20 & 52.00 & 25.93 & 0 & 18.10 & 31.50 & 17.87 & 0 & 45.90 & 68.70 & 47.57 & 0 & 37.80 & 45.40 & 37.55 & 0 \\
  \bottomrule
  \end{tabular}
  }
  \caption{Comparison results of explicit privacy preserving.}
  \label{explicit}
\end{table*}


In actual scenarios, users sometimes know the privacy-related information in the sentences to be translated, but sometimes they may not. In this paper, we classify task scenarios into two categories based on whether the privacy-related terms in the sentences are known in advance.
\paragraph{Scenario 1: Explicit Privacy-Preserving Machine Translation}
In this case, all privacy-related terms in sentences are known, and we do not need to identify which words belong to privacy-related terms.

\paragraph{Scenario 2: Implicit Privacy-Preserving Machine Translation}
In this case, we have no information about which words are privacy-related information, so we need to identify the privacy-related terms first.
    
To more fully measure the performance of our approach, we test it in both scenarios. In reality, the first scenario is more general as users usually have certain requirements on the information they do not want to disclose, and the second scenario requires an additional module for identifying privacy information on our approach. Since different people have different definitions of privacy information, we use Texsmart~\cite{texsmart2021} to identify certain type of entities and treat them as privacy-related terms for the sake of the reproducibility of the experimental results.

Moreover, we find that the translation of time and quantity entities is more difficult than the translation of the other three types of entities. To reflect the difficulty of translating time and quantity entities, we divided the results corresponding to one translation model into two types: those containing only the \textbf{person}, \textbf{places} and \textbf{organization} (3 types), and those containing all five categories (5 types).

\subsection{Results on Explicit Privacy-Preserving Scenario}

\subsubsection{General Domain}
The results of our proposed method on WMT18 and WMT20 human-annotated test datasets are shown in the table.~\ref{explicit}. The second line shows the translation performance of the Transformer model on standard bilingual test datasets. Placeholder means which kind of placeholder is utilized to replace the private information in the input sentence. The format of the other tables is similar to that.

In this table, we compare three types of translation model. For their translation performance evaluated by three general metrics (SacreBLEU, chrF and BLEU), the standard transformer that takes sentences whose privacy information is replaced by dictionary-based placeholder as input is worse than Dict-Trans and Tag-Trans. In addition, Dict-Transformer and Tag-Trans show comparable translation performance.

For privacy-preserving performance, we could see that in the explicit privacy-preserving scenario, all privacy information are replaced by these placeholders. Therefore, no privacy information has been leaked.

\subsubsection{Finance Domain}
In this dataset, the number of time and quantity entities is larger than the number of other types of entity. Therefore, we could find that the performance of the translation models in 3 types is significantly better than that in 5 types.

\subsection{Results on Implicit Privacy-Preserving Scenario}
\label{Implicit_Results}

\begin{table*}[bt]
  \centering
  \resizebox{\linewidth}{!}{
  \begin{tabular}{cccccccccccccccccccccccccccccc}
  \toprule
  \multicolumn{2}{c}{\multirow{3}*{\shortstack{Placeholder}}} & \multicolumn{2}{c}{\multirow{3}*{\shortstack{Translation Model}}} & \multicolumn{8}{c}{WMT18}  & \multicolumn{8}{c}{WMT20}  & \multicolumn{8}{c}{Finance} \\
   
   \cmidrule(lr){5-12}\cmidrule(lr){13-20}\cmidrule(lr){21-28}
   \multicolumn{4}{c}{} & \multicolumn{4}{c}{\ZhEn} & \multicolumn{4}{c}{\EnZh}  & \multicolumn{4}{c}{\ZhEn} & \multicolumn{4}{c}{\EnZh}  & \multicolumn{4}{c}{\ZhEn} & \multicolumn{4}{c}{\EnZh} \\
   \cmidrule(lr){5-8}\cmidrule(lr){9-12} \cmidrule(lr){13-16} \cmidrule(lr){17-20} \cmidrule(lr){21-24}\cmidrule(lr){25-28}
   \multicolumn{4}{c}{} & SacreBLEU  & chrF & BLEU  & PER & SacreBLEU  & chrF & BLEU & PER & SacreBLEU  & chrF & BLEU & PER & SacreBLEU  & chrF & BLEU & PER & SacreBLEU  & chrF & BLEU & PER & SacreBLEU  & chrF & BLEU & PER \\
   \midrule
   \multicolumn{2}{c}{-} & \multicolumn{2}{c}{Trans} & 22.80 & 51.40 & 22.61 & - & 18.60 & 29.50 & 18.18 & - & 27.00 & 55.50 & 27.90 & - & 19.20 & 33.10 & 19.06 & - & 47.40 & 71.20 & 48.19 & - & 39.60 & 47.30 & 38.57 & - \\
   \midrule
   \multicolumn{2}{c}{Entity} & \multicolumn{2}{c}{Trans (3 types)} &
   19.80 & 44.20 & 19.72 & 0.1851 & 15.70 & 25.90 & 15.59 & 0.2477 & 24.90 & 49.40 & 24.83 & 0.1859 & 16.90 & 27.30 & 16.81 & 0.2272 & 47.20 & 67.50 & 47.14 & 0.1587 & 37.20 & 44.60 & 37.11 & 0.2004 \\
   \midrule
   \multicolumn{2}{c}{Entity} & \multicolumn{2}{c}{Trans (5 types)} & 
   19.40 & 43.30 & 19.27 & 0.1913 & 15.10 & 24.80 & 15.01 & 0.2579 & 24.30 & 48.20 & 24.22 & 0.1913 & 16.20 & 26.10 & 16.14 & 0.2355 & 46.10 & 66.00 & 46.03 & 0.1641 & 36.00 & 43.10 & 35.91 & 0.2066 \\
   \midrule
   \multicolumn{2}{c}{Dictionary} & \multicolumn{2}{c}{Dict-Trans (3 types)} &
   21.50 & 48.60 & 21.41 & 0.1845 & 17.50 & 28.10 & 17.23 & 0.2485 & 26.80 & 53.00 & 26.68 & 0.1865 & 18.90 & 33.10 & 18.86 & 0.2280 & 49.10 & 70.80 & 48.99 & 0.1595 & 39.00 & 46.80 & 38.92 & 0.2010 \\
   \midrule
   \multicolumn{2}{c}{Dictionary} & \multicolumn{2}{c}{Dict-Trans (5 types)} &
   21.10 & 47.90 & 21.04 & 0.1925 & 16.90 & 27.10 & 16.86 & 0.2585 & 26.30 & 52.10 & 26.19 & 0.1920 & 18.20 & 31.70 & 18.11 & 0.2355 & 47.90 & 69.20 & 47.82 & 0.1645 & 38.10 & 45.40 & 37.79 & 0.2075 \\
   \midrule
   \multicolumn{2}{c}{Tag} & \multicolumn{2}{c}{Tag-Trans (3 types)} &
   21.60 & 48.60 & 21.46 & 0.1875 & 17.80 & 28.10 & 17.31 & 0.2395 & 26.60 & 52.80 & 26.55 & 0.1775 & 18.80 & 32.90 & 18.78 & 0.2200 & 48.90 & 70.60 & 48.78 & 0.1485 & 39.00 & 46.60 & 38.84 & 0.1900 \\
   \midrule
   \multicolumn{2}{c}{Tag} & \multicolumn{2}{c}{Tag-Trans (5 types)} &
   20.90 & 47.20 & 20.88 & 0.1835 & 17.00 & 27.30 & 16.77 & 0.2450 & 26.20 & 52.00 & 26.11 & 0.1835 & 18.00 & 31.50 & 17.97 & 0.2275 & 47.80 & 69.00 & 47.71 & 0.1535 & 37.90 & 45.20 & 37.67 & 0.1950 \\

  \bottomrule
  \end{tabular}
  }
  \caption{Comparison results of implicit privacy preserving.}
  \label{implicit}
\end{table*}

The current scenario requires additional consideration of the identification of private terms in sentences compared to the previous scenario. Since the focus of this paper is not on the identification of privacy information, here we only use the entities identified by the named entity recognition tool as privacy-related terms. The results in this scenario are listed in the table.~\ref{implicit}. 

From the two tables, we can see a significant increase in PER on all datasets compared to the previous scenario, which is because the named entity recognition tool did not identify all the privacy. However, this also leads to an improvement in the quality of the translations obtained. In addition, we could observe that the PER values in the \EnZh direction are significantly higher than the PER values in the \ZhEn direction. This is because the named entity recognition tool that we use performs worse in recognizing English entities. For the performance of different placeholders, we could find that the tag-based placeholder outperforms the dictionary-based placeholder in terms of privacy protection in implicit privacy-preserving situations.

Moreover, although the PER values are around 0.2 in the current scenario, this is acceptable for overall privacy protection. This means that only one of the five privacy terms is not protected, which is not very serious for overall privacy protection. More importantly, this problem is not caused by our privacy-preserving framework, but is related to the effectiveness of automatic privacy identification tools. In our future work, we will explore the use of efficient automatic privacy identification tools in conjunction with the current framework.

\subsection{Comparison with Other Methods}

To validate the effectiveness of our method in the domain of privacy protection, we compared it with two classic privacy-preserving techniques in natural language processing: one based on local differential privacy (named Embedding-LDP~\cite{embedding-ldp}) and another that combines secure multiparty computation and full homomorphic encryption (named Power-Softmax~\cite{Zimerman2024PowerSoftmaxTS}). Both methods provide absolute protection of the input sentences (that is, PER = 0), but require additional computational overhead and may degrade the quality of the translation model.

Furthermore, with the increasing availability of computational resources and advancements in large language models, deploying large models locally for translation has also emerged as an option for privacy-preserving machine translation. To compare the performance of our framework with that of a large model deployed locally, we conducted experiments using the non-thinking mode of Qwen3-8B~\cite{Yang2025Qwen3TR}.

In the following, we present a comparative analysis of the inference performance of our method and the aforementioned approaches on the WMT18 and Finance datasets in a scenario involving five types of privacy entity. The comparison covers three aspects: the computational latency on the user's side, the translation performance, and the PER values. The user latency is the cost time of the user-side computation module to process the full test dataset.
 
Since the aforementioned methods lack dedicated implementations for Transformer models, we developed the relevant code ourselves.

For Embedding-LDP, we consider the computation of the token embedding layer, the embedding noise injection module, and the result denoising module as client-side computations, while all other computations are treated as server-side. To ensure consistency with the original paper and guarantee practical effectiveness, we trained a Transformer model as the denoising module for the generation task.

For Power-Softmax, we reused the parameters of a standard pre-trained Transformer and conducted the Polynomial Transformer Construction described in the paper to obtain an encryption-enhanced Transformer. During inference, we consider the Attention component as a server-side computation and the remaining parts as client-side computation.

Based on experimental results in the table.~\ref{comparison}, it can be observed that although Embedding-LDP and Power-Softmax offer strong privacy protection capabilities, they require modifications to the Transformer's computational structure, impose significant computational overhead on the client side, and lead to a substantial degradation in machine translation performance. In contrast, our tag-based method incurs minimal client-side overhead while maintaining high machine translation quality.

The Qwen3-8B model demonstrates exceptionally strong translation capabilities and provides absolute privacy protection through local deployment. However, this also comes with significantly higher computational resource requirements. In our experiments using four V100 GPUs, due to memory constraints, we set the batch size to 8 and the maximum generation length to 512. As a result, the total processing time reaches approximately 2,000 seconds, far exceeding that of other methods.

\begin{table*}[bt]
  \centering
  \resizebox{1.0\linewidth}{!}{
  \begin{tabular}{cccccccccccccccccccccccc}
  \toprule
   \multicolumn{2}{c}{\multirow{3}*{\shortstack{Placeholder}}} & \multicolumn{2}{c}{\multirow{3}*{\shortstack{Translation Model}}} & \multicolumn{10}{c}{WMT18}  & \multicolumn{10}{c}{Finance}\\
   
   \cmidrule(lr){5-14}\cmidrule(lr){15-24}
   \multicolumn{4}{c}{} & \multicolumn{5}{c}{\ZhEn} & \multicolumn{5}{c}{\EnZh}  & \multicolumn{5}{c}{\ZhEn} & \multicolumn{5}{c}{\EnZh} \\
   \cmidrule(lr){5-9}\cmidrule(lr){10-14} \cmidrule(lr){15-19} \cmidrule(lr){20-24}
   \multicolumn{4}{c}{} & SacreBLEU  & chrF & BLEU & PER & User Latency & SacreBLEU  & chrF & BLEU & PER & User Latency & SacreBLEU  & chrF & BLEU & PER & User Latency & SacreBLEU  & chrF & BLEU & PER  & User Latency \\
   \midrule
   \multicolumn{2}{c}{-} & \multicolumn{2}{c}{Trans} & 22.80 & 51.40 & 22.61 & - & - & 18.60 & 29.50 & 18.18 & - & - & 47.40 & 71.20 & 48.19 & - & - & 39.60 & 47.30 & 38.57 & - & - \\
   \midrule
   \multicolumn{2}{c}{Dictionary} & \multicolumn{2}{c}{Dict-Trans (5 types)} &
    21.10 & 47.90 & 21.04 & 0.1925 & 28.9s & 16.90 & 27.10 & 16.86 & 0.2585 & 24.5s & 47.90 & 69.20 & 47.82 & 0.1645 & 26.1s & 38.10 & 45.40 & 37.79 & 0.2075 & 21.9s \\

   \multicolumn{2}{c}{Tag} & \multicolumn{2}{c}{Tag-Trans (5 types)} &
   20.90 & 47.20 & 20.88 & 0.1835 & 21.6s & 17.00 & 27.30 & 16.77 & 0.2450 & 19.7s & 47.80 & 69.00 & 47.71 & 0.1535 & 19.3s & 37.90 & 45.20 & 37.67 & 0.1950 & 17.8s \\
   \midrule
   \multicolumn{2}{c}{-} & \multicolumn{2}{c}{PowerSoftmax} &
   17.40 & 39.10 & 17.31 & 0 & 58.2s & 14.20 & 23.40 & 14.12 & 0 & 47.7s & 43.40 & 64.80 & 43.36 & 0 & 50.2s & 35.20 & 42.60 & 35.14 & 0 & 44.9s \\

   \multicolumn{2}{c}{-} & \multicolumn{2}{c}{Embedding LDP} & 18.50 & 41.60 & 18.47 & 0 & 32.3s & 15.70 & 25.80 & 15.62 & 0 & 26.5s & 44.60 & 66.50 & 44.59 & 0 & 29.7s & 35.70 & 42.80 & 35.63 & 0 & 24.3s \\
    \midrule
   \multicolumn{2}{c}{-} & \multicolumn{2}{c}{Qwen3-8B} & 23.50 & 54.20 & 23.47 & 0 & 1802.0s & 19.70 & 31.20 & 19.68 & 0 & 1744.4s & 52.40 & 74.10 & 52.40 & 0 & 2247.1s & 41.40 & 49.10 & 41.33 & 0 & 2193.6s \\

  \bottomrule
  \end{tabular}
  }
  \caption{Comparison results of our framework and other baseline methods in implicit privacy preserving. We conduct experiments on WMT18 and Finance 5 types datasets. The user latency is the cost time of user side's computation module for processing the full test dataset.}
  \label{comparison}
\end{table*}

   

\begin{table*}[bt]
  \centering
  \resizebox{1.0\linewidth}{!}{
  \begin{tabular}{cccccccccccccccccccccc}
  \toprule
    \multicolumn{2}{c}{\multirow{3}*{\shortstack{Translation Model}}} & \multicolumn{6}{c}{WMT18}  & \multicolumn{6}{c}{WMT20} & \multicolumn{6}{c}{Finance} \\
   \cmidrule(lr){3-8}\cmidrule(lr){9-14}\cmidrule(lr){15-20}
   \multicolumn{2}{c}{} & \multicolumn{3}{c}{\ZhEn} & \multicolumn{3}{c}{\EnZh}  & \multicolumn{3}{c}{\ZhEn} & \multicolumn{3}{c}{\EnZh}  & \multicolumn{3}{c}{\ZhEn} & \multicolumn{3}{c}{\EnZh} \\
   \cmidrule(lr){3-5}\cmidrule(lr){6-8}\cmidrule(lr){9-11}\cmidrule(lr){12-14}\cmidrule(lr){15-17}\cmidrule(lr){18-20}
   \multicolumn{2}{c}{} & SacreBLEU  & chrF & BLEU & SacreBLEU  & chrF & BLEU & SacreBLEU  & chrF & BLEU & SacreBLEU  & chrF & BLEU &  SacreBLEU  & chrF & BLEU &  SacreBLEU  & chrF & BLEU  \\
   \midrule
    \multicolumn{2}{c}{dict-trans (3 types)} &
    21.30 & 48.50 & 21.20 & 17.10 & 27.90 & 17.08 & 26.60 & 53.00 & 26.50 & 18.80 & 33.10 & 18.77 & 48.90 & 70.80 & 48.87 & 38.80 & 46.80 & 38.78 \\
    \multicolumn{2}{c}{trans (3 types)} & 20.40 & 46.20 & 20.29 & 16.30 & 26.40 & 16.23 & 25.50 & 50.50 & 25.40 & 17.90 & 30.90 & 17.81 & 47.80 & 68.70 & 47.73 & 37.70 & 45.10 & 37.62 \\
    \midrule
    \multicolumn{2}{c}{dict-trans(5 types)}	& 21.00 & 47.40 & 20.81 & 16.50 & 27.10 & 16.45 & 26.10 & 52.10 & 26.01 & 18.00 & 31.70 & 17.98 & 47.70 & 69.20 & 47.69 & 37.70 & 45.40 & 37.63 \\
    \multicolumn{2}{c}{trans (5 types)} & 20.00 & 45.40 & 19.96 & 15.60 & 25.80 & 15.59 & 25.20 & 50.20 & 25.14 & 17.20 & 29.80 & 17.13 & 46.90 & 67.50 & 46.82 & 37.00 & 44.30 & 36.90 \\
  \bottomrule
  \end{tabular}
  }
  \caption{Translation Performance of Trans and Dict-Trans models used to process dictionary-based placeholder.}
  \label{trans_dict_placeholder}
\end{table*}

\label{DA_section}

\begin{table*}
  \centering
  \resizebox{\linewidth}{!}{
  \begin{tabular}{cccccccccccccccccccc}
  \toprule
   \multicolumn{2}{c}{\multirow{3}*{\shortstack{Placeholder}}} & \multicolumn{2}{c}{\multirow{3}*{\shortstack{Translation Model}}} & \multicolumn{8}{c}{Implicit Privacy Preserving}  & \multicolumn{8}{c}{Explicit Privacy Preserving} \\
   
   \cmidrule(lr){5-12}\cmidrule(lr){13-20}
   \multicolumn{4}{c}{} & \multicolumn{4}{c}{\ZhEn} & \multicolumn{4}{c}{\EnZh}  & \multicolumn{4}{c}{\ZhEn} & \multicolumn{4}{c}{\EnZh} \\
   \cmidrule(lr){5-8}\cmidrule(lr){9-12} \cmidrule(lr){13-16} \cmidrule(lr){17-20}
   \multicolumn{4}{c}{} & SacreBLEU & chrF & BLEU & PER & SacreBLEU & chrF & BLEU & PER & SacreBLEU & chrF & BLEU & PER & SacreBLEU & chrF & BLEU & PER \\
   
   \midrule
   
   \multicolumn{2}{c}{-} & \multicolumn{2}{c}{Trans} & 27.00 & 55.50 & 27.90 & - & 19.20 & 33.10 & 19.06 & - & 27.00 & 55.50 & 27.90 & - & 19.20 & 33.10 & 19.06 & - \\
   \midrule
   \multicolumn{2}{c}{Dictionary} & \multicolumn{2}{c}{Dict-Trans w/o DA (3 types)} &
   26.20 & 52.00 & 26.13 & 0.1892 & 18.50 & 32.20 & 18.39 & 0.2298 & 26.10 & 51.90 & 26.03 & 0 & 18.30 & 32.10 & 18.26 & 0 \\
   \midrule
   \multicolumn{2}{c}{Dictionary} & \multicolumn{2}{c}{Dict-Trans w/o DA (5 types)} & 25.60 & 51.10 & 25.54 & 0.1958 & 17.70 & 30.80 & 17.65 & 0.2382 & 25.50 & 51.00 & 25.48 & 0 & 17.50 & 30.70 & 17.42 & 0 \\
   \midrule
   \multicolumn{2}{c}{Dictionary} & \multicolumn{2}{c}{Dict-Trans (3 types)} & 26.80 & 53.00 & 26.68 & 0.1865 & 18.90 & 33.10 & 18.86 & 0.2280	& 25.70 & 52.90 & 26.50 & 0 & 18.90 & 33.10 & 18.77 & 0 \\
   \midrule
   \multicolumn{2}{c}{Dictionary} & \multicolumn{2}{c}{Dict-Trans (5 types)} & 26.30 & 52.10 & 26.19 & 0.1920 & 18.20 & 31.70 & 18.11 & 0.2355	& 25.30 & 52.00 & 26.01 & 0 & 18.20 & 31.80 & 17.98 & 0 \\
   \midrule
   \multicolumn{2}{c}{Tag} & \multicolumn{2}{c}{Tag-Trans w/o DA (3 types)} & 26.10 & 51.90 & 26.02 & 0.1802 & 18.30 & 32.10 & 18.24 & 0.2135 & 25.60 & 51.80 & 25.91 & 0 & 18.20 & 32.00 & 18.18 & 0 \\
    \midrule
   \multicolumn{2}{c}{Tag} & \multicolumn{2}{c}{Tag-Trans w/o DA (5 types)} & 25.60 & 51.10 & 25.53 & 0.1878 & 17.40 & 30.60 & 17.34 & 0.2316 & 25.20 & 51.10 & 25.52 & 0 & 17.50 & 30.60 & 17.41 & 0 \\
   \midrule
   \multicolumn{2}{c}{Tag} & \multicolumn{2}{c}{Tag-Trans (3 types)} & 26.60 & 52.80 & 26.55 & 0.1775 & 18.80 & 32.90 & 18.78 & 0.2200 & 25.60 & 52.50 & 26.34 & 0 & 18.80 & 32.90 & 18.67 & 0 \\
    \midrule
   \multicolumn{2}{c}{Tag} & \multicolumn{2}{c}{Tag-Trans (5 types)} & 26.20 & 52.00 & 26.11 & 0.1835 & 18.00 & 31.50 & 17.97 & 0.2275 & 25.20 & 52.00 & 25.93 & 0 & 18.10 & 31.50 & 17.87 & 0 \\

  \bottomrule
  \end{tabular}
  }
  \caption{The performance of translation models trained with and w/o data augmentation on the WMT20 dataset.}
  \label{DA}
\end{table*}

\subsection{Adaptive Training's Necessity}
We introduced adaptive training when using dictionary-based placeholders to improve the final translation performance. The need for adaptive training arises because the dictionary-based method replaces private entities in the input text with entries selected from a small, fixed-size translation lexicon. This increases the frequency of entities from the dictionary, altering thereby the text distribution. Hence, adaptive training is necessary to adapt to this shift.

\begin{table*}
    \centering
    \resizebox{0.8\linewidth}{!}{
    \begin{tabular}{cccccc}
    \toprule
      & Method &	WMT18 \ZhEn &	WMT18 \EnZh &	Finance \ZhEn &	Finance \EnZh \\
      \midrule
        \multirow{6}*{\shortstack{Stanford CoreNLP}} & Entity (3 types) &	0.1766 &	0.2508	& 0.1672 &	0.2114 \\
        & Entity (3 types)	& 0.1822	& 0.2611 &	0.1699 &	0.2154 \\
        & Dict (3 types) &	0.1905 &	0.2498 &	0.1623 &	0.2087 \\
        & Dict (5 types)	& 0.1935 &	0.2377 &	0.1677	& 0.2132 \\
        & Tag (3 types) &	0.1912 &	0.2345 &	0.1493 &	0.1934 \\
        & Tag (5 types)	& 0.1866 &	0.2465	& 0.1567	& 0.1978 \\
        \hline
        \multirow{6}*{\shortstack{Microsoft Recognizers-Text}} & Entity (3 types)	& 0.1811	& 0.2453	& 0.1598	& 0.2054 \\
        & Entity (5 types)	& 0.1907	& 0.2512	& 0.1662	& 0.2078 \\
        & Dict (3 types)	& 0.1831	& 0.2454	& 0.1647	& 0.2056 \\
        & Dict (5 types)	& 0.1888	& 0.2519	& 0.1687	& 0.2106 \\
        & Tag (3 types)	& 0.1871	& 0.2366	& 0.1476	& 0.1921 \\
        & Tag (5 types)	& 0.1845	& 0.2407	& 0.1531	& 0.1978 \\
        \hline
        \multirow{6}*{\shortstack{Multiple Recognizers Fusion}} & Entity (3 types)	& 0.1633	& 0.2092	& 0.1434	& 0.1914 \\
        & Entity (5 types)	& 0.1688	& 0.2147	& 0.1478	& 0.1898 \\
        & Dict (3 types)	& 0.1645	& 0.2112	& 0.1496	& 0.1901 \\
        & Dict (5 types)	& 0.1679	& 0.2156	& 0.1477	& 0.1975 \\
        & Tag (3 types)	& 0.1605	& 0.2047	& 0.1389	& 0.1806 \\
        & Tag (5 types)	& 0.1597	& 0.2082	& 0.1424	& 0.1821 \\
        \hline
        \multirow{6}*{\shortstack{Roberta-Recognizer}} & Entity (3 types)	& 0.1323	& 0.2006	& 0.1367	& 0.1869 \\
        & Entity (5 types)	& 0.1339	& 0.2076	& 0.1389	& 0.1807 \\
        & Dict (3 types)	& 0.1376	& 0.2021	& 0.1383	& 0.1825 \\
        & Dict (5 types)	& 0.1356	& 0.2034	& 0.1335	& 0.1881 \\
        & Tag (3 types)	& 0.1411	& 0.1925	& 0.1310	& 0.1797 \\
        & Tag (5 types)	& 0.1424	& 0.1976	& 0.1372	& 0.1768 \\
        \bottomrule
    \end{tabular}}
    \caption{PER Results on WMT18 and financial datasets with different Named Entity Recognizers. Multiple Recognizer Fusion means that we use the merge results from TexSmart, Stanford CoreNLP, MicroSoft Recognizers-Text.}
    \label{NER_tool}
\end{table*}

To support this view, we conduct experiments to compare the translation performance of trans and dict-trans models when used to process text that contains dictionary-based placeholders. The BLEU results are shown in the table.~\ref{trans_dict_placeholder}. It can be observed that, without adaptive training, the SacreBLEU score decreases by 0.7 to 1.1, the chrF score decreases by 1.1 to 2.5, and the BLEU score decreases by 0.8 to 1.2 points. This demonstrates the necessity of incorporating adaptive training when using the dictionary-based placeholder.

\subsection{The Effects of Data Augmentation}

To reveal the effects of data augmentation, we list the results of translation models trained with/without data augmentation on the WMT20 dataset in the table.~\ref{DA}. According to this table, we could find that translation models trained with the data augmentation method outperform translation models without training with the data augmentation method in both translation quality and privacy protection. In addition, the results show more gains in the quality of model translation from data enhancement in implicit privacy-preserving scenarios. This is because less privacy information is replaced in this scenario, and the input and data-enhanced datasets are closer together.

\subsection{Entity Recognizer's Impact on PER}
To demonstrate the impact of different NER tools on the final PER metric, we conduct experiments of implicit privacy preserving using additional recognizers, including Stanford Core NLP~\cite{manning-EtAl:2014:P14-5} and Microsoft Recognizers-Text~\cite{soft:recognizers-text}, Multiple-Recognizer Fusion, Roberta-Recognizer on WMT18 and finance datasets. The results are shown in the table.~\ref{NER_tool}. The Multiple-Recognizer Fusion means that we use the union of recognition results of Texsmart, Stanford Core NLP, Microsoft Recognizers-Text as the final results. Roberta recognizer is a roberta-base model~\cite{liu2019robertarobustlyoptimizedbert} trained to recognize the named entity, and the training data is collected using the above three tools to process the original training datasets.

Based on the results, we can observe that the performance of individual NER tools (Stanford CoreNLP, Microsoft Recognizers-Text) shows little difference. However, employing multiple NER tools combined with training a dedicated neural network recognition model can effectively improve the PER metric of our framework.

\subsection{NER Error's Impact on Translation and Privacy Protection Performance}

To investigate the impact of NER errors on both privacy leakage and translation performance, we divide the test cases into two categories: those without NER errors and those with NER errors, and evaluated the model's performance on these two scenarios on the finance 5 types dataset. Detailed results can be found in the table.~\ref{ner_error}.

Based on the results in the table, it can be observed that in the absence of NER recognition errors, our framework achieves absolute privacy protection. Furthermore, since the primary type of NER errors is missed entity recognition, the translation performance of the model shows a significant improvement in cases with such errors.

   

\begin{table}[bt]
  \centering
  \resizebox{1.0\linewidth}{!}{
  \begin{tabular}{cccccccccccc}
  \toprule
   \multicolumn{2}{c}{\multirow{3}*{\shortstack{Translation Mode}}} & \multicolumn{2}{c}{\multirow{3}*{\shortstack{Case Type}}} & \multicolumn{8}{c}{Finance 5 types}  \\
   
   \cmidrule(lr){5-12}
   \multicolumn{4}{c}{} & \multicolumn{4}{c}{\ZhEn} & \multicolumn{4}{c}{\EnZh}   \\
   \cmidrule(lr){5-8}\cmidrule(lr){9-12} 
   \multicolumn{4}{c}{} & SacreBLEU & chrF & BLEU & PER & SacreBLEU & chrF & BLEU & PER   \\
   \midrule
   \multicolumn{2}{c}{Trans} & \multicolumn{2}{c}{-} & 47.40 & 71.20 & 48.19 & - & 39.60 & 47.30 & 38.57 & - \\
   \midrule
   \multicolumn{2}{c}{Dict-Trans} & \multicolumn{2}{c}{No NER Error} &46.90 & 69.20 & 47.82 & 0 & 37.80 & 45.40 & 37.79 & 0    \\
   
   \multicolumn{2}{c}{Dict-Trans} & \multicolumn{2}{c}{NER Error} &47.20 & 69.50 & 48.07 & 0.2533 & 38.20 & 45.70 & 38.14 & 0.2756    \\ 
   \midrule
   \multicolumn{2}{c}{Tag-Trans} & \multicolumn{2}{c}{No NER Error} & 46.80 & 69.00 & 47.71 & 0 & 37.70 & 45.20 & 37.67 & 0    \\
   \multicolumn{2}{c}{Tag-Trans} & \multicolumn{2}{c}{NER Error} & 47.30 & 69.40 & 48.11 & 0.2316 & 38.30 & 45.80 & 38.24 & 0.2776 \\
  \bottomrule
  \end{tabular}
  }
  \caption{NER Error's Influence to the Translation Performance and PER on the Fintech (5 types) dataset}
  \label{ner_error}
\end{table}

\begin{table*}[bt]
  \centering
  \resizebox{\linewidth}{!}{
  \begin{tabular}{cccccccccccccccccccc}
  \toprule
   \multicolumn{2}{c}{\multirow{3}*{\shortstack{Placeholder}}} & \multicolumn{2}{c}{\multirow{3}*{\shortstack{Translation Model + Recognizer}}} & \multicolumn{8}{c}{Finance 10 types}  & \multicolumn{8}{c}{Finance CP} \\
   
   \cmidrule(lr){5-12}\cmidrule(lr){13-20}
   \multicolumn{4}{c}{} & \multicolumn{4}{c}{\ZhEn} & \multicolumn{4}{c}{\EnZh}  & \multicolumn{4}{c}{\ZhEn} & \multicolumn{4}{c}{\EnZh} \\
   \cmidrule(lr){5-8}\cmidrule(lr){9-12} \cmidrule(lr){13-16} \cmidrule(lr){17-20} 
   \multicolumn{4}{c}{} & SacreBLEU & chrF & BLEU & PER & SacreBLEU & chrF & BLEU & PER & SacreBLEU & chrF & BLEU & PER & SacreBLEU & chrF & BLEU & PER  \\
   \midrule
   \multicolumn{2}{c}{-} & \multicolumn{2}{c}{Trans} & 47.40 & 71.20 & 48.19 & - & 39.60 & 47.30 & 38.57 & - & 47.40 & 71.20 & 48.19 & - & 39.60 & 47.30 & 38.57 & -  \\
   \midrule
   \multicolumn{2}{c}{Entity} & \multicolumn{2}{c}{Trans (5 types)} &
   46.10 & 66.00 & 46.03 & 0.2240 & 36.00 & 43.10 & 35.91 & 0.2566 & 46.10 & 66.00 & 46.03 & 0.2557 & 36.00 & 43.10 & 35.91 & 0.2769    \\
   \midrule
   \multicolumn{2}{c}{Dictionary} & \multicolumn{2}{c}{Dict-Trans (5 types)+ Texsmart (5 types)} & 47.90 & 69.20 & 47.82 & 0.2235 & 37.80 & 45.40 & 37.79 & 0.2601 & 47.90 & 69.20 & 47.82 & 0.2686 & 37.80 & 45.40 & 37.79 & 0.2777   \\
   \multicolumn{2}{c}{Tag} & \multicolumn{2}{c}{Tag-Trans (5 types)+ Texsmart (5 types)} &47.80 & 69.00 & 47.71 & 0.2145 & 37.70 & 45.20 & 37.67 & 0.2513 & 47.80 & 69.00 & 47.71 & 0.2632 & 37.70 & 45.20 & 37.67 & 0.2769  \\
   \midrule
   \multicolumn{2}{c}{Dictionary} & \multicolumn{2}{c}{Dict-Trans (5 types) + Texsmart (10 types)} &
   45.40 & 65.20 & 45.36 & 0.1833 & 35.40 & 42.40 & 35.33 & 0.2305 & - & - & - & - & - & - & - & -  \\
   \multicolumn{2}{c}{Tag} & \multicolumn{2}{c}{Tag-Trans (5 types) + Texsmart (10 types)} &45.50 & 65.30 & 45.49 & 0.1879 & 35.80 & 42.90 & 35.75 & 0.2344 & - & - & - & - & - & - & - & - \\
   \midrule
   \multicolumn{2}{c}{Dictionary} & \multicolumn{2}{c}{Dict-Trans (5 types) + Contextual Privacy Recognizer} &- & - & - & - & - & - & - & - & 44.20 & 63.50 & 44.15 & 0.2071 & 33.30 & 39.90 & 33.24 & 0.2145  \\
   \multicolumn{2}{c}{Tag} & \multicolumn{2}{c}{Tag-Trans (5 types) + Contextual Privacy Recognizer} & - & - & - & - & - & - & - & - & 44.00 & 63.30 & 44.01 & 0.2023 & 33.40 & 40.00 & 33.31 & 0.2201 \\

  \bottomrule
  \end{tabular}
  }
  \caption{Results of implicit privacy preserving on extended datasets. We still use the Texsmart as the 10 types recognizer and train a Roberta-base model as the contexual privacy recognizer.}
  \label{extended_datasets}
\end{table*}

\subsection{Extension of Privacy Scope}

The primary objective of our study is to construct a relatively universal academic benchmark. We selected these five types of entities because they account for a high proportion (approximately 80\%) of entities related to privacy, as confirmed by our experimental validation. Based on these five categories, the dataset allows us to create a broadly applicable privacy entity dataset with relatively low manual annotation costs. However, this approach limits the practical application of our datasets.

To mitigate this issue, we expand the scope of privacy content in the test set through two methods: first, by adding five new types of entity [Company, Product, Work, Brand, Position] and second, by annotating a contextual privacy dataset (as mentioned by reviewer 3) not limited to specific entities. Ultimately, with the assistance of both human annotators and large language models, we annotated our financial domain dataset, resulting in two new datasets: one expanded from the original five entity types to include [Company, Product, Work, Brand, Position], which we refer to as Finance 10 types, and another encompassing contextual privacy, named Finance CP, for testing purposes. The annotation criterion for contextual privacy is that if a text span relates to a private entity through semantic association, it is considered contextual privacy. Furthermore, we used GPT4 to annotate contextual privacy texts in finance training data and trained a contextual privacy recognition based on the roberta-base model to identify such instances.

We evaluate the model trained on the five original entity types in these two datasets. and the results are presented in the table.~\ref{extended_datasets}. It can be observed that the performance of our framework slightly decreases when handling more entity types and contextual privacy, compared to processing only five categories of private entities. However, it remains within an acceptable range.

\begin{CJK*}{UTF8}{gbsn}
\subsection{Case Study}

  \begin{table*}
  \centering
  \resizebox{\linewidth}{!}{
  \begin{tabular}{ccc}
  \toprule
   \multicolumn{2}{c}{\shortstack{Source}} & \textcolor{red}{联合国 国际 原子能 总署} （ \textcolor{red}{IAEA} ） \textcolor{red}{26} 日 公布 报告 指出 ， \textcolor{red}{伊朗} 已 开始 使用 先进 的 离心机  \\ & & 加速 生产 浓缩铀 ， 进一步 违反 \textcolor{red}{2015} 年 与 世界大国 达成 的 核 协议 。 \\
   \midrule
   
   \multicolumn{2}{c}{Reference} &  \textcolor{red}{The United Nations International Atomic Energy Agency} ( \textcolor{red}{IAEA}) released a report on the \textcolor{red}{26th} stating that \textcolor{red}{Iran} \\ & &  has started to use advanced centrifuges to accelerate the production of enriched uranium , further violating the nuclear agreement reached with world powers in \textcolor{red}{2015} . \\
   \midrule
   \multicolumn{2}{c}{Standard Transformer Generation} & The \textcolor{red}{United Nations International Atomic Energy Agency} ( \textcolor{red}{IAEA} ) released a report \textcolor{red}{Tuesday} that said \textcolor{red}{Iran}  \\ & &  has begun to use advanced centrifuges to accelerate the production of enriched uranium , further violating its nuclear deal with the world 's major powers in \textcolor{red}{2015} . \\
   \midrule
   \multicolumn{2}{c}{Tag-based Privacy Replacement} &   \textcolor{red}{ $\langle PINFO0 \rangle$} （  \textcolor{red}{$\langle PINFO1 \rangle$} ）  \textcolor{red}{5} 日 公布 报告 指出 ，  \textcolor{red}{$\langle PINFO2 \rangle$} 已 开始 使用 先进 的 离心机 \\ & & 加速 生产 浓缩铀 ， 进一步 违反  \textcolor{red}{2005} 年 与 世界大国 达成 的 核 协议 。   \\
   \midrule
   \multicolumn{2}{c}{Tag-Trans Generation} & The \textcolor{red}{$\langle PINFO0 \rangle$} ( \textcolor{red}{$\langle PINFO1 \rangle$} ) reported \textcolor{red}{yesterday} that \textcolor{red}{$\langle PINFO2 \rangle$} had begun to use advanced centrifuges  \\ &  & to accelerate the production of enriched uranium , in further violation of the \textcolor{red}{2005} nuclear agreement with world powers . \\
   \midrule
   \multicolumn{2}{c}{Final Translation} & The \textcolor{red}{UN 's IAEA} ( \textcolor{red}{IAEA} ) reported \textcolor{red}{26} that \textcolor{red}{Iran} had begun to use advanced centrifuges \\ &  & to accelerate the production of enriched uranium , in further violation of the \textcolor{red}{2015} nuclear agreement with world powers .\\

  \bottomrule
  \end{tabular}
  }
  \caption{A case of translation with Tag-Trans. The red parts in listed sentences are privacy information.}
  \label{Tag Case Study}
\end{table*}

\begin{table*}
  \centering
  \resizebox{\linewidth}{!}{
  \begin{tabular}{ccc}
  \toprule
   \multicolumn{2}{c}{\shortstack{Source}} & \textcolor{red}{联合国 国际 原子能 总署} （ \textcolor{red}{IAEA} ） \textcolor{red}{26} 日 公布 报告 指出 ， \textcolor{red}{伊朗} 已 开始 使用 先进 的 离心机  \\ & & 加速 生产 浓缩铀 ， 进一步 违反 \textcolor{red}{2015} 年 与 世界大国 达成 的 核 协议 。 \\
   \midrule
   
   \multicolumn{2}{c}{Reference} &  \textcolor{red}{The United Nations International Atomic Energy Agency} ( \textcolor{red}{IAEA}) released a report on the \textcolor{red}{26th} stating that \textcolor{red}{Iran} \\ & &  has started to use advanced centrifuges to accelerate the production of enriched uranium , further violating the nuclear agreement reached with world powers in \textcolor{red}{2015} . \\
   \midrule
   \multicolumn{2}{c}{Standard Transformer Generation} & The \textcolor{red}{United Nations International Atomic Energy Agency} ( \textcolor{red}{IAEA} ) released a report \textcolor{red}{Tuesday} that said \textcolor{red}{Iran}  \\ & &  has begun to use advanced centrifuges to accelerate the production of enriched uranium , further violating its nuclear deal with the world 's major powers in \textcolor{red}{2015} . \\
   \midrule
   \multicolumn{2}{c}{Dictionary-based Privacy Replacement} & \textcolor{red}{联合国} （ \textcolor{red}{安全理事会} ） \textcolor{red}{5} 日 公布 报告 指出 ， \textcolor{red}{中国} 已 开始 使用 先进 的 离心机  \\ & & 加速 生产 浓缩铀 ， 进一步 违反 \textcolor{red}{2005} 年 与 世界大国 达成 的 核 协议 。 \\
   \midrule
   \multicolumn{2}{c}{Dict-Trans Generation} & The \textcolor{red}{United Nations} ( \textcolor{red}{Security Council} ) released its report on \textcolor{red}{Tuesday} that \textcolor{red}{china} \\ & & has begun to use advanced centrifuges to accelerate the production of enriched uranium , further violating the \textcolor{red}{2005} nuclear deal with world powers .\\
   \midrule
   \multicolumn{2}{c}{Final Translation} & the \textcolor{red}{UN 's IAEA} ( \textcolor{red}{IAEA} ) released its report on \textcolor{red}{26} that \\ & &  \textcolor{red}{Iran} has begun to use advanced centrifuges to accelerate the production of enriched uranium , further violating the \textcolor{red}{2015} nuclear deal with world powers . \\


  \bottomrule
  \end{tabular}
  }
  \caption{A case of translation with Dict-Trans. The red parts in listed sentences are privacy information.}
  \label{dict Case Study}
\end{table*}

\begin{table*}
  \centering
  \resizebox{\linewidth}{!}{
  \begin{tabular}{ccc}
   \toprule
   \multicolumn{2}{c}{\shortstack{Source Language Sentence}} & 年内\textcolor{red}{上市公司 (PE 1)} 回购 规模 超 \textcolor{red}{1000 亿元 (PE 2)}  创 历史 新高 - \textcolor{red}{新华网 (PE 3)}  。 \\
   \midrule

   \multicolumn{2}{c}{Target Language Reference Sentence} &  \textcolor{red}{Listed company (PE 1)} repurchase scale hits record high of over \textcolor{red}{100 billion yuan (PE 2)} during the year - \textcolor{red}{Xinhuanet (PE 3)}. \\
   \midrule
   \multicolumn{2}{c}{Dictionary-based Privacy Replacement} & 年内\textcolor{red}{联合国 (D-P 1)} 回购 规模 超 \textcolor{red}{130 元 (D-P 2)}  创 历史 新高 - \textcolor{red}{安全 理事会 (D-P 3)}  。 \\
   \midrule
   \multicolumn{2}{c}{Dict-Trans Generation} & \textcolor{red}{United Nations(D-P 1)} Buy Back to Top \textcolor{red}{130(D-P 2)} in Year - \textcolor{red}{Security Council(D-P 3)} \\
   \midrule
   \multicolumn{2}{c}{Final Translation} & \textcolor{red}{Listed companies(PE 1)} buy back to top \textcolor{red}{ 1000(PE 2)} in year - \textcolor{red}{Xinhua(PE 3)} .\\


  \bottomrule
  \end{tabular}
  }
  \caption{A failure case of translation with Dict-Trans. The red parts in above sentences are privacy information. Specifically, PE x means the x-th privacy entity and D-P x means Dict-Placeholder for x-the privacy entity.}
  \label{Failure Case Study}
\end{table*}

In order to investigate the strengths and weaknesses of our proposed method more deeply and concretely, we did case studies to see which parts our method can handle well and which parts it is more prone to error.

The cases of translation with two translation models are shown in the table.~\ref{Tag Case Study} and the table.~\ref{dict Case Study}, respectively. The case is a translation from Zh to En. We list the source sentence, reference sentence, the generation of the transformer model with standard training and inputs, the sentence with privacy-related terms replaced, our translation model's generation, and the final translation after decoding the privacy information in the generation of our model. The red parts in listed sentences are privacy-related terms or their corresponding placeholders.

According to the two tables, we could find that most of the non-privacy parts are translated well although we remove some privacy information in the source sentence. However, we still could observe that our methods tend to output a shorter translation compared with the standard Transformer model when translating  "进一步 违反 2015 年 与 世界大国 达成 的 核 协议 。(further violating the nuclear agreement reached with world powers in 2015)".

For the translation of private information, we can find that the placeholders of private entities are well preserved in the dictionary-based generation. In our experiments, about 96\%-98\% of the placeholders for private information are preserved, which means that our models learn this mapping well. The errors often happen in the process of looking up a phrase translation table. When the privacy entity is not in this table, it is hard to translate this entity well. What's more, in our case, the final translation of  "联合国 国际 原子能 总署 (The United Nations International Atomic Energy Agency)" is correct when it is judged by a human. However, when we use the BLEU score to evaluate this sentence, the BLEU score drops due to the difference in the form of expression between this entity's translation and the standard translation.

Moreover, we provide a failure case in the table.~\ref{Failure Case Study}. Our discussion here primarily focuses on the types of translation errors, as privacy leaks are mainly caused by the NER tools' failure to identify private entities and are minimally influenced by our framework. The types of translation errors include entity translation error, and over-replacement for privacy entity. According to the table.~\ref{Failure Case Study}, the correction translation of the privacy entity "新华网 (Xinhuanet)" is "Xinhuanet" but the translation word table gives another result "Xinhua". This is an entity translation error, which is related to the quality of translation word table. For the privacy entity 1000 亿元 (100 billion yuan)", it is replaced to "130 元 (130 yuan)". This replacement leads to the missing of "亿 (1/10 billion)", which means 1/10 billion in Chinese, which is an over-replacement error, which is due to inaccurate recognition of the recognizer and may be overcome by using a better recognizer.

\end{CJK*}

\section{Limitation}
Our limitations cover the scope of privacy, the NER tools, and the constraints compared to locally deployed LLMs.

First, defining privacy solely based on named entities is an oversimplification, as many types of privacy, such as the contextual privacy you mentioned later, extend beyond entities. To address this, we rely on specialized privacy detection tools, including directly employing NER tools and trained neural networks. However, these tools still struggle to ensure absolute privacy protection. Lastly, with increasing accessibility of deployment frameworks and computational resources, the cost of locally deploying large models is expected to decrease. This will likely make local deployment a more viable option for users, prompting us to reconsider the future direction of our framework in the era of large language models.

\section{Conclusion}
In order to protect users' privacy information when they use external translation services, in this paper, we propose a new machine translation task: privacy-preserving machine translation. To facilitate the research on this task, we provide test datasets with manual annotations on privacy-related words and introduce evaluation metrics to measure the effects of privacy leakage. Then we provide a set of benchmark models based on the split-translate-merge framework for this task. Experiments show that these methods are promising to maintain the translation quality and protect privacy information and this further demonstrates that it is possible to build privacy-preserving NMT.

\newpage

\section*{Acknowledgments}
This work was supported in part by the Research Grants Council of the Hong Kong SAR under Grant GRF 11217823, 11216225, and Collaborative Research Fund C1042-23GF, the National Natural Science Foundation of China under Grant 62371411, InnoHK initiative, the Government of the HKSAR,Laboratory for AI-Powered Financial Technologies.


\bibliography{custom}
\bibliographystyle{IEEEtran}

\newpage

 




\vfill

\end{document}